\documentclass[10pt,twocolumn,letterpaper]{article}

\pdfoutput=1

\usepackage{cvpr}
\usepackage{times}
\usepackage{epsfig}
\usepackage{graphicx}
\usepackage{amsmath}
\usepackage{amssymb}

\usepackage[caption=false,font=footnotesize]{subfig}
\usepackage{subfloat}

\usepackage{subfig}
\usepackage{epstopdf}
\usepackage{multirow}
\usepackage{enumitem}

\usepackage{color}

\newcommand\fisheyewidth{0.7}

\usepackage[breaklinks=true,bookmarks=false]{hyperref}

\cvprfinalcopy 

\def\cvprPaperID{1624} 

\begin{document}

\title{Cost-based Feature Transfer for Vehicle Occupant Classification}

\author{Toby Perrett, Majid Mirmehdi\\
University of Bristol\\
{\tt\small tp8961@bristol.ac.uk, m.mirmehdi@bristol.ac.uk}
\and
Eduardo Dias\\
Jaguar Land Rover\\
{\tt\small edias@jaguarlandrover.com}
}

\maketitle

\begin{abstract}

Knowledge of human presence and interaction in a vehicle is of growing interest to vehicle manufacturers for design and safety purposes.  We present a framework to perform the tasks of occupant detection and occupant classification for automatic child locks and airbag suppression.  It operates for all passenger seats, using a single overhead camera.  A transfer learning technique is introduced to make full use of training data from all seats whilst still maintaining some control over the bias, necessary for a system designed to penalize certain misclassifications more than others.  An evaluation is performed on a challenging dataset with both weighted and unweighted classifiers, demonstrating the effectiveness of the transfer process.

\end{abstract}

\section{Introduction}

One of the main areas in which computer vision can aid vehicle design is improving safety for occupants.  Examples of this include lane tracking \cite{Bottazzi2013,Hanwell2012},  collision prediction systems \cite{SarahBonninThomasH.WeisswangeFranzKummert2014,Monwar2013,Sivaraman2013}, and alertness monitoring for the driver \cite{Vicente2015,Garcia2012}.  One such area that has received particular {research interest  is automatic occupant classification \cite{Huang2012,Huang2011,Gao2010,Goktuk2005}}.  This has traditionally been for the purpose of airbag suppression, where it is sometimes safer to not deploy an airbag in a collision if the occupant is a small child \cite{Glass2000,Member2001}. 

{Non-vision sensors have been used for the tasks of occupant detection \cite{George2010,mehney2000vehicle} and classification \cite{wallace2003vehicle}, but there are a number of incentives to replace these with a camera based system.  These include the possibility of introducing additional functionality without any additional hardware (for example, determining if an occupant is using a control surface \cite{Cheng2010}), and reducing the cost to the vehicle manufacturer.}

Designing a computer vision system to work reliably in a safety-critical automotive environment presents a number of challenges.  These include a large variety of lighting conditions (including over exposure in some cases), skin tones, clothing, postures and occupant behavior.  The bias of the system should also be considered - is it better to incorrectly identify an adult as a child or vice versa, for example?  Finally, any evaluation should be robust enough to highlight the system's shortcomings, should any exist.

{In this paper, a method is proposed that can handle the tasks of occupant detection and occupant classification, that functions in all passenger seats using a single camera.  An occupant detection system would allow a vehicle to automatically turn on infotainment systems and seatbelt warnings, for example.  The two use cases of occupant classification we address here are airbag suppression and automatic child locks.  Currently, these two features, which have different age cutoffs and intra-class costs, have to be switched off and on manually, leading to the possibility of human error.} 
To address the issues raised above, it is important to make full use of the available training data from every seat.  Thus, a transfer learning technique is introduced to allow a classifier for one seat to use training information from others whilst still maintaining some control over {the bias generated by the transfer process}.  More specifically, Histogram of Orientated Gradient (HOG) and binned motion statistics are extracted from training images and used as features.  Joint Discriminant Analysis (JDA) is applied to find a space of reduced dimensionality shared by both the source and target features.  Source features are then transferred to bring them closer to the target feature distributions whilst being attracted to classes with a lower misclassification cost.  Finally, these features are used to train a weighted Support Vector Machine (SVM) classifier, which can then classify unseen images from the target seat.

\begin{figure*}
\begin{centering}
    \subfloat[An empty cabin.]{\includegraphics[width=\fisheyewidth\columnwidth]{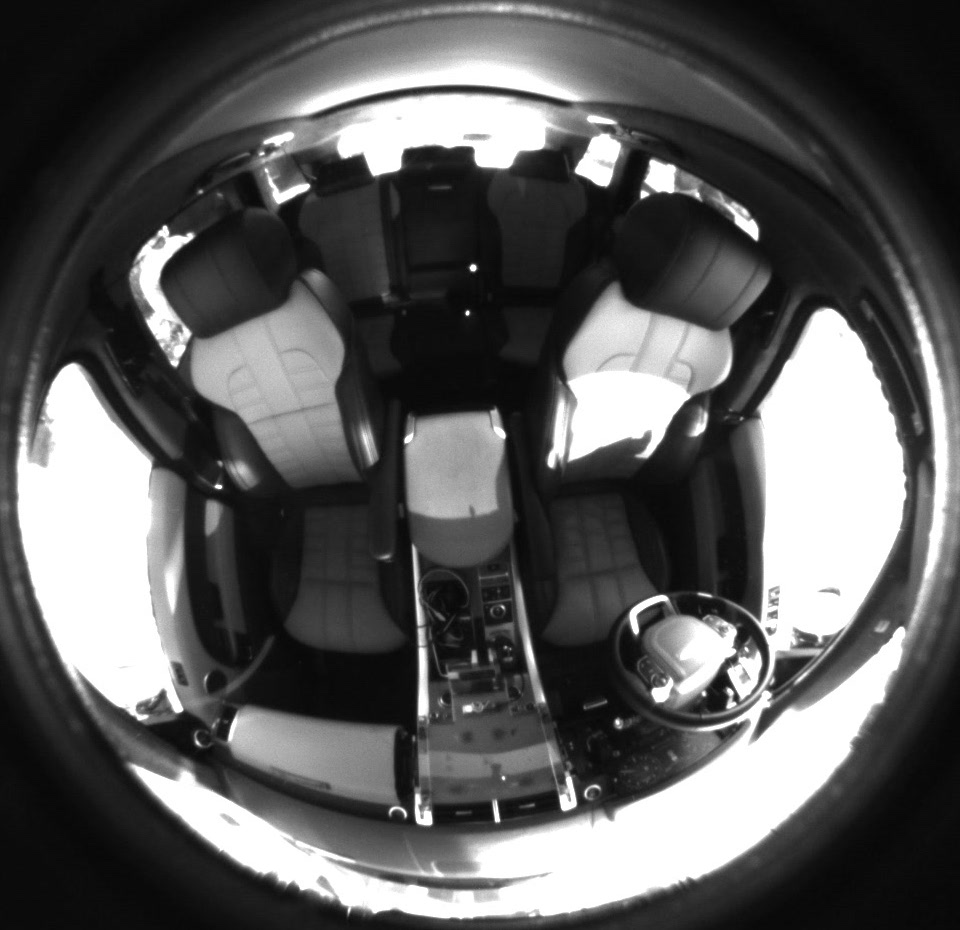}} \hspace{2mm}
    \subfloat[Three children.  The front passenger is not seated correctly, and the rear passengers are putting seatbelts on.]{\includegraphics[width=\fisheyewidth\columnwidth]{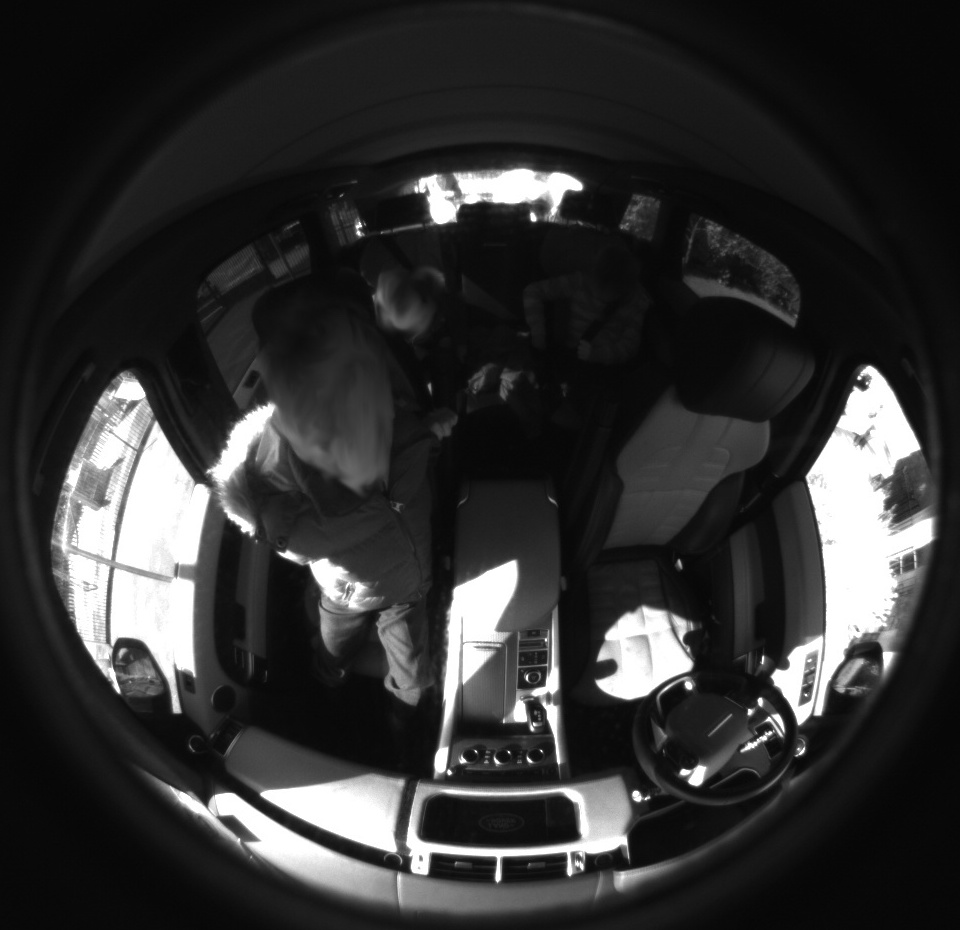}} \hspace{2mm}
    \subfloat[Two children and one adult.  Some overexposure is present, particularly in the front passenger region.]{\includegraphics[width=\fisheyewidth\columnwidth]{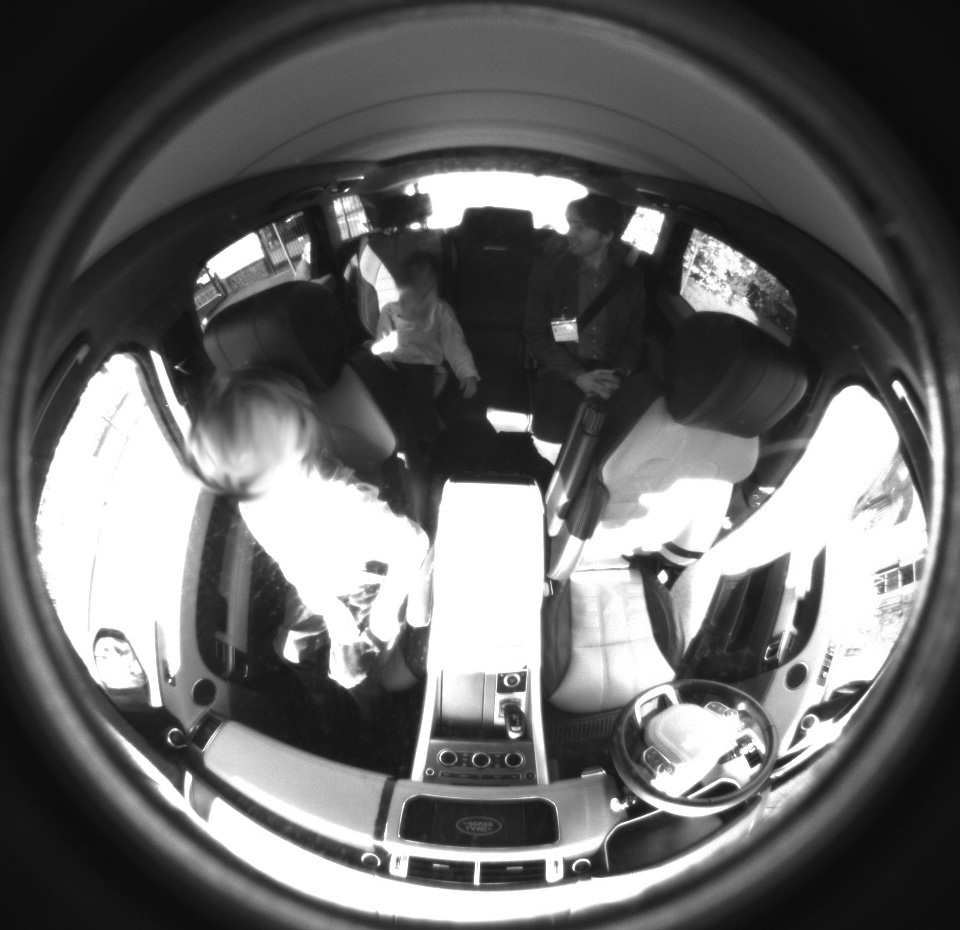}}
\end{centering}
\caption{Some example images from our dataset with child faces blurred.}
\label{fig:dataset_examples}
\end{figure*}

{To evaluate the proposed method, a dataset was collected featuring 40 adults and 60 children. {To the best of our knowledge} this is the first time such a large {dataset, including children,}  has been used to evaluate an occupant classification method using computer vision.}

{Our contributions in this work are threefold.  Perhaps the most important is the proposed cost-based transfer learning method.  This is the first time intra-class costs have been addressed from a transfer learning perspective, and applied to a weighted classification problem.  Secondly, computer vision has not previously been used to investigate automatic child locks, which is presented here in the same framework as an automatic airbag suppression method. Finally, this is also the first time a single sensor has been used to perform the occupant classification task for multiple seats simultaneously}. Whilst this paper attempts to solve occupant detection and classification tasks, the transfer learning method has the potential for wider use in computer vision.   It would be most applicable in cases where learning from multiple sources is desirable but not currently used due to a high cost of misclassification for some classes. For example, a road sign classification system could incorporate additional sources of training data, while taking into account that it is  worse to misclassify a 20mph road sign as a 60mph than it is to classify a 50mph as a 30mph.

Next, in Section \ref{sec:related_work}, a review of occupant classification literature and related techniques is given.  Section \ref{sec:dataset} introduces the dataset and Section \ref{sec:method} presents the proposed method, which is evaluated and discussed in Section \ref{sec:experiments}.  Finally, we conclude and suggest future work in Section \ref{sec:conclusion}.

\section{Related work }\label{sec:related_work}

Camera-based occupant classification systems are not normally used in production vehicles, so we will begin by giving an overview of different sensors and associated methods and assessing their suitability for our task.  We will then justify our choice of pursuing a computer vision solution, and introduce related computer vision occupant classification literature.

The majority of occupant detection systems rely on weight sensors in the seat \cite{mehney2000vehicle}, and are used to warn when an occupant is not wearing a seatbelt.  Arrays of weight sensors have been suggested as a way to perform occupant classification by monitoring the pressure on different parts of the seat \cite{wallace2003vehicle}, but there are issues which are yet to be addressed.  For example, the weight of a child seat (some weigh over 10kg) or an overweight child could could result in the confusion of the small and large child classes, causing an airbag deployment instead of a suppression.  Additionally, an automotive-grade weight sensor costs around three times that of a single camera, so there are cost savings to switching to a computer vision approach.  
Ultrasound \cite{seip2002linear} and inductive \cite{George2010} sensors have also been suggested for the occupant detection task, but while detection with them is possible, they have some shortcomings when applied to the classification task. A child in a booster seat could be confused for an adult when using overhead ultrasound sensors as their heads would be similar heights, and inductive sensors provide no way to distinguish between occupant types.

Several works have been published using multiple camera setups to obtain depth information \cite{Kong2004,Cheng2004} to then help with a foreground segmentation stage.  Similarly, time of flight cameras have also been used for depth information from frontal \cite{Goktuk2005} and side-on \cite{Alefs2008} views.  
One potential issue with using depth sensors to monitor all the vehicle occupants is that they have a limited  depth range in which they operate correctly, and also a field of view that makes observing more than one occupant difficult.  A cheap, single camera with a wide field of view lens (e.g. see Figure \ref{fig:dataset_examples}) would be the ideal choice of sensor from a manufacturer's perspective\footnote{This work is supported and guided by Jaguar Land Rover.} as one could be used to observe the whole cabin, assuming it could demonstrate acceptable accuracy.  

One of the major problems when using cameras in an automotive environment is the variability of illumination conditions.  Proposed solutions to this problem are to use high dynamic range (HDR) imagery \cite{Koch2002} or infra-red illumination at night time and cameras that operate in this frequency range \cite{Farmer2007}.

Most vision-based occupant classification systems, such as \cite{Farmer2003,Kong2004,Goktuk2005,Zhang2005,Devarakota2007,Alefs2008,Gao2010,Huang2011},  tend to follow a standard classification pipeline, usually consisting of a feature extraction stage, dimensionality reduction stage, and classification stage.  
However, there are issues that are yet to be addressed in the literature.  These include how a system should work on more than one seating position at a time can most effectively use all the training data available, and how methods perform on a dataset with real children (not just dolls and blankets \cite{Goktuk2005,Farmer2007,Alefs2008}) and with a robust evaluation (\ie not leave one out cross-validation \cite{Farmer2003} or subjects appearing in both training and test sets \cite{Huang2010}).

Intuitively, occupancy detection and classification seems like an ideal problem for improvement via some transfer of knowledge.  We will  classify occupants in four separate seats, and while these seating areas will not look identical, there should be some kind of correlation, for example, in feature space between a child in a front seat and a child in a rear seat.  This will allow us to leverage more training data, as training data for each seat can be used to help in the training process in all the others if necessary.

Traditionally, transfer learning has focussed on one of three approaches: unsupervised, transductive, or inductive \cite{Pan2010a}.  Unsupervised learning (such as \cite{Baktashmotlagh2014})
 takes place when no class labels are known for either the source or target domains, so is not applicable here.  Inductive transfer occurs when some labelled examples are known for both the source and the target domains \cite{Wang2014,Shao2012}.  Some works contain a large number of source labels and a small number of target labels, but in the proposed dataset there will also be a large number of target labels which enables a better understanding of the target distribution. This makes it more suitable for the adaptation of transductive \cite{Campos2013,Rohrbach2013} techniques (which are used when no target labels are known, but tend to rely on information about the shape of the whole source and target distributions) on a class by class basis.  An interesting transductive example is that by Farajidavar \etal \cite{Farajidavar2014}, who introduced a method for fine-tuning feature projections in a subspace shared by both the source and target domains.  Whilst labels were not available for the target data, it was assumed that the entire target distribution (\ie all the features for the target test data) was known as a prior.   This contrasts with the data we have, where we have some labelled examples from source and target domains, but their entire distributions are unknown.

\section{Dataset}\label{sec:dataset}

Whilst previous methods have used blankets and dolls to simulate children, these cannot provide the necessary variability to adequately evaluate a system that is critical in nature.  As such, for this study we recruited 60 children between the ages of 3 and 9, along with 40 adults.  A high variability in the data was ensured by asking the participants to spend some time behaving normally, and some time ``acting up'' for the camera.  There are a large number of clothing types, skin tones, lighting conditions, postures and behaviors in the footage.

A single near infra-red (NIR) camera with a 190$^\circ$ field of view lens was mounted just above the rear view mirror, so that all the seats were fully visible apart from the rear nearside and offside seats, which were mostly visible.  {An NIR camera was chosen as it is more robust to the wide variety of illumination conditions found in vehicles, including the ability to operate at night. 

The first 30-60 seconds were filmed after the occupants entered the vehicle.  Participants were seated in different configurations (\eg adult in front passenger seat, child in booster seat in rear nearside, empty rear middle and rear offisde), with each participant taking part in around three different sequences.  In total, there are 122 sequences {each containing footage of 4 seats, some with occupants and some empty}. Examples are given in Figure \ref{fig:dataset_examples}.

\section{Proposed method} \label{sec:method}

In the proposed approach, after some preprocessing, features are extracted from each frame and put through a dimensionality reduction stage.  An initial location in the feature space, shared by the source and target domains, for each feature is calculated, followed by a procedure to adjust these locations based on the cost of misclassification.  The final stage is to train a classifier that takes these features as an input and outputs the predicted class of the occupant.

\subsection{Data preparation, feature extraction and dimensionality reduction}
	
The first stage of the processing  is to correct the fisheye distortion of the $190^{\circ}$ FOV lens by following the calibration procedure of Scaramuzza \etal \cite{Scaramuzza2006}.
	
Various features have been used in the literature for occupancy detection, including Haar responses \cite{Huang2011}, 
edge features \cite{Zhang2005}, 
and shape statistics \cite{Farmer2007}, 
amongst others.  We trialled a number of these, but found using HOG features, concatenated with optical flow \cite{Brox2004} based motion descriptors (the mean flow magnitude and orientation from the previous 5 frames were taken from each bin), produced the best results.

{Once image features are extracted, dimensionality reduction needs to be performed before they can be passed to a classifier.  Standard dimensionality reduction techniques such as Principal Component Analysis (PCA) or more recent methods \cite{Coifman2005} would not be applicable for this task, as the distributions of the source and target features are not necessarily the same.
A better option is to use a method that is aware of different source and target distributions.  Dimensionality can be reduced while simultaneously transforming the feature distributions of source and target features so they more closely match.
A commonly used technique is to exploit the Maximum Mean Discrepancy (MMD) measure \cite{Gretton2012}, which gives a measurement of how different two distributions are.  Examples can be found in \cite{Kim2013} and \cite{Pan2008}, amongst others.  Long \etal \cite{Long2013} introduced Joint Discriminant Analysis (JDA), where they expanded on the MMD measure by allowing it to determine the differences in marginal and conditional distributions of the features.  Figure \ref{fig:dim_red} shows example reduced source and target feature distributions generated by JDA.}

\begin{figure}
\centering
    \subfloat[Source - front passenger.]{\includegraphics[width=0.95\columnwidth]{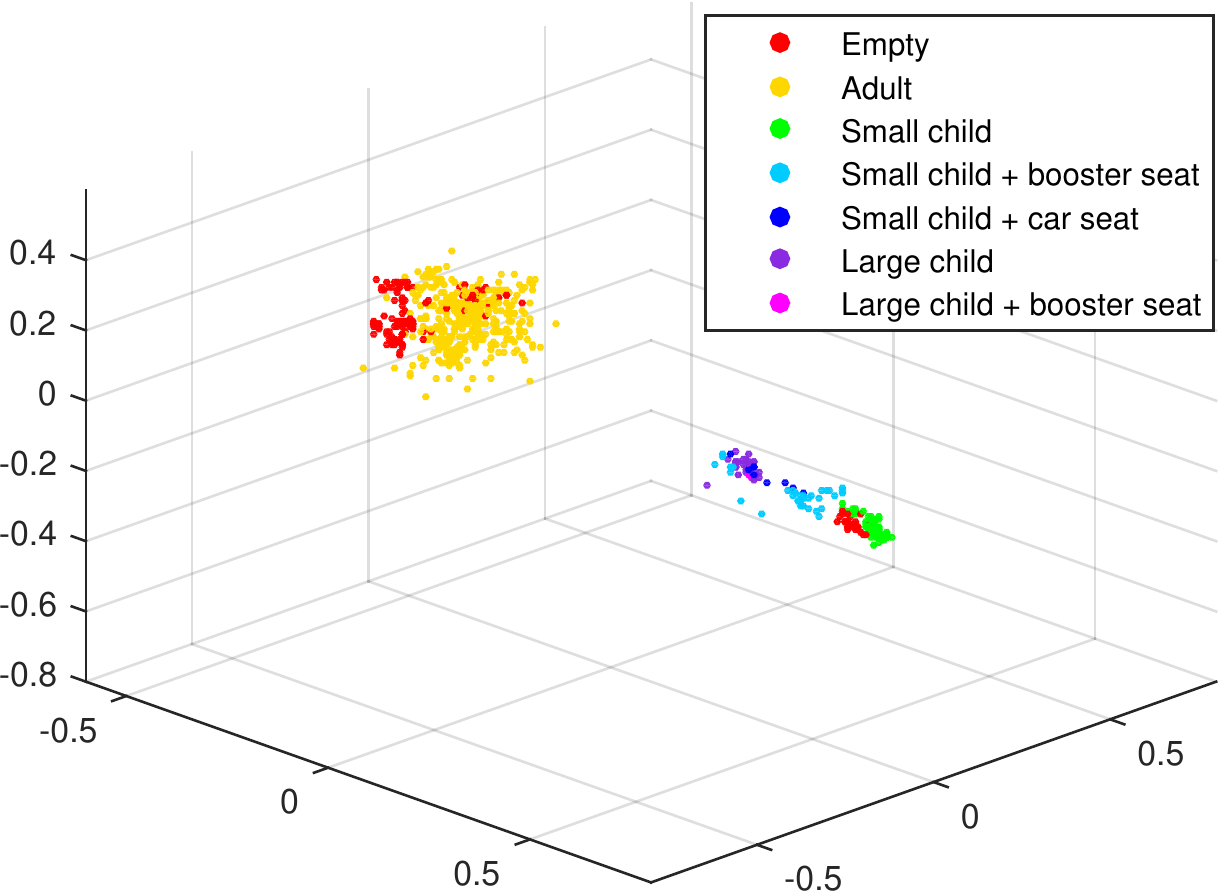}}   \\
    \subfloat[Target - rear offside passenger.]{\includegraphics[width=0.95\columnwidth]{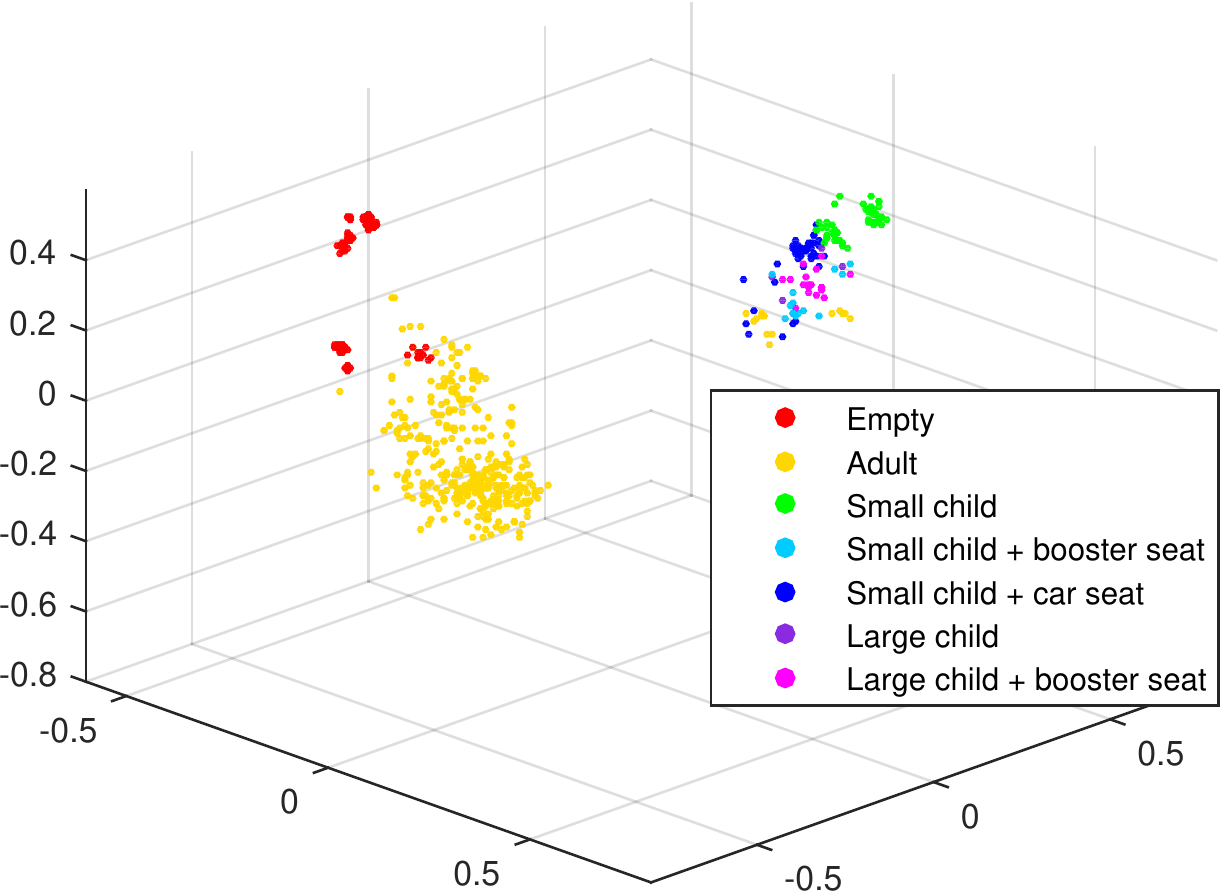}}
\caption{Example source and target distributions in the shared reduced space.  Note, only the three most significant dimensions from the JDA process are shown, while our classifier uses 50 dimensions (where better separation between the classes are achieved).}
\label{fig:dim_red}
\end{figure}

\subsection{Feature transfer via weighted Gaussians} \label{sec:gaussians}
		
		Once the initial embeddings of the source features in the shared reduced space have been found, we would like to take into account the way classes are arranged in this shared space and adjust each feature's location accordingly.
		
		The first step is to model the location and distribution of each source class, which is done by fitting a mixture of Gaussians to each class separately.  Here, we take $N$ classes with $G$ Gaussians per class, and say the $j$th Gaussian belonging to the $i$th class has the following mean, covariance and weighting parameters 		
		\begin{equation} \label{eq:gauss_src_params}
			\lambda_{ij}^{src} = \left\lbrace \mu_{ij}^{src}, \Sigma_{ij}^{src}, w_{ij}^{src}  \right\rbrace.
		\end{equation}
		  Each source feature $f$ can then be represented by a matrix $F^{src}$ containing its relative likelihood with respect to each Gaussian on a class by class basis.  We write the $i$th row and $j$th column as		
		\begin{equation} \label{eq:F_src1}
			F_{ij}^{src} = \frac{w_{ij}^{src}P(f | \lambda_{ij}^{src})}{\sum_j w_{ij}^{src}P(f | \lambda_{ij}^{src}) }.
		\end{equation}
		The same process is then performed for $f$ with respect to the target classes, which gives		
		\begin{equation} \label{eq:gauss_tar_params}
			\lambda_{ij}^{tar} = \left\lbrace \mu_{ij}^{tar}, \Sigma_{ij}^{tar}, w_{ij}^{tar}  \right\rbrace
		\end{equation}
and
		\begin{equation} \label{eq:F_tar1}
			F_{ij}^{tar} = \frac{w_{ij}^{tar}P(f | \lambda_{ij}^{tar})}{\sum_j w_{ij}^{tar}P(f | \lambda_{ij}^{tar}) }.
		\end{equation}		
Given a set of all $S$ source features $\left\lbrace f_1,...,f_S \right\rbrace$ with class labels $\left\lbrace l_1,...,l_S \right\rbrace$, we would like to find new features $\left\lbrace \tilde f_1,...,\tilde f_S \right\rbrace$ with the same labels such that their probabilities of lying in the target class distribution are similar to how they lie in the source class distribution.  This is achieved by defining a function
		\begin{equation} \label{eq:min1}
			T(\tilde f_k) = \sum_i \sum_j D(F_{kij}^{src}, \tilde F_{kij}^{tar}),
		\end{equation}
 where $D$ is a similarity measure (\eg the norm function), which is then minimized for every pair $f_k^{src}$ and $\tilde f_k^{tar}$.  For this case, a high penalty should be imposed for large differences between $F_k^{src}$ and $F_k^{tar}$, whereas smaller differences will be less important to correct.  It will also be desirable for a large number of non-zero differences to contribute, so the $L^3$ norm is chosen.  Note how $D$ may have to be chosen differently if applied to different tasks.
		
Minimising (\ref{eq:min1}) is necessary in the case where all classes are treated equally, \ie where the metric used to evaluate the system is the true positive rate.  However, in our occupant detection and classification system, it is desirable to have some control over the bias of this feature transfer procedure.  For example, when deciding whether to activate a child lock, it does not matter if the ``empty'' class is classified as ``adult'' or ``child'', but a manufacturer would be concerned about confusion of the ``adult'' and ``child'' classes.  Moreover, they may wish to bias the system by penalising a misclassification of a child as an adult more than a misclassification of an adult as a child for safety reasons, or vice versa so as not to cause embarrassment for the vehicle owner.
		
Equation (\ref{eq:min1}) can also take into account this desired bias if (\ref{eq:F_tar1}) is modified.  First, an $N$ by $N$ cost matrix $\Phi$ is defined, where the entry $\Phi(u,v)$ denotes the cost of misclassifying class $u$ as class $v$.  The condition $\forall u,  \forall v,  0 \leq \Phi(u,v) \leq 1 $ is enforced.  Next, a monotonically increasing function $\Psi$ is chosen, such that $\forall x \in \left[ 0 , 1 \right] , \Psi(x) > 0  $.  Equation (\ref{eq:F_tar1}) then becomes	
		\begin{equation} \label{eq:F_tar2}
			F_{ij}^{tar} = \frac{ \Psi(\Phi_{il_k}) w_{ij}^{tar}P(f | \lambda_{ij}^{tar})}{\sum_j  \Psi(\Phi_{il_k}) w_{ij}^{tar}P(f | \lambda_{ij}^{tar}) }.
		\end{equation}		
The choice of $\Psi$ and $\Phi$ is very important to the overall effectiveness of the proposed method.  The more ``aggressive'' $\Psi$ is (\ie the harder it pulls features towards the centre of their target class distribution and pushes them away from target class distributions which have a higher cost), the lower the probability of the transferred features causing an undesirable misclassification.  However, increasing this aggression means that the transferred features are less likely to provide useful additional information to the classifier. Conversely, a weaker $\Psi$ can influence a classifier's predictions more, but at the possible cost of more undesirable misclassifications.  A similar situation occurs when choosing $\Phi$.  Costs of either 0 or 1 could be chosen, reducing the classification to a two class problem, but this would mean less information about the overall shape of the target distribution would be transferred.  {We determined the best choices for $\Psi$ and $\Phi$ empirically for our experiments.  As an example, for the classes labelled as Empty (E), Adult (A), Small child (S) and Large child (L) for the occupant detection case, $\Psi$ was chosen as the exponential function and $\Phi$ as:}
\begin{equation}
\Phi = \bordermatrix{~ & \mbox{E} & \mbox{A} & \mbox{S} & \mbox{L} \cr
                  \mbox{E} & 0 & 0.4 & 0.2 & 0.3 \cr
                  \mbox{A} & 0.4 & 0 & 0.2 & 0.1 \cr
                  \mbox{S} & 0.2 & 0.4 & 0 & 0.1 \cr
                  \mbox{L} & 0.3 & 0.1 & 0.1 & 0 \cr}.
\end{equation}

\subsection{Gradient descent procedure}
	
	Now that we have an objective function to minimize, we can proceed by gradient descent (we use the method from Lagarias \etal \cite{Lagarias1998}).  However, an important issue that has to be considered is where this process should start from.  As we are dealing with high dimensional data (in practice, 50 dimensions are retained after the dimensionality reduction stage) with the potential for a large number of local minima, the final state of a gradient descent process is likely to be sensitive to its initial state.  As mentioned in Section \ref{sec:related_work}, Farajidavar \etal \cite{Farajidavar2014} proposed a method for transferring features when target labels are not known.  Their method relied on an initial embedding based on the overall shapes of the source and target data as the target labels were not available.  This initial embedding could be used here, but as we have access to target labels it is appropriate to incorporate this additional information to enhance the initial embedding.  We thus proceed by adopting the approach in \cite{Farajidavar2014}, but using it one class at a time.
	
	For each source feature $f_k$, an initial embedding $\tilde f_k$ is taken, where $\tilde f_k = f_k + \tau u_k$, where $\tau$ is a large scaling factor.    The quantity $u_k$ is obtained by taking the Gaussians fitted to the target classes with parameters $\lambda_{l_k,j}^{tar}$ for $j=0,...,G$.  Then we set
		\begin{equation} \label{eq:uk}
			u_{kd} = \frac{1}{S \sqrt{w_{l_kj}^{tar}}}\sum_k P(\lambda_{l_kj}^{tar} | f_k) \frac{f_{dk} - \mu_{d l_k j}^{tar}}{\sigma_{d l_k j}^{tar}}.
		\end{equation}
This quantity can be easily computed by taking the mean component of the Fischer vector from \cite{Perronnin2006}.
	
\subsection{Classification}
For the classification stage, an SVM is used {with a radial basis function (RBF) kernel}.  When training a classifier to evaluate the performance over all classes with no weightings provided (in Section \ref{sec:overall_classification}), a grid search over the parameters for the LIBSVM implementation is conducted \cite{CC01a}.
When training and testing weightings for specific tasks (in Section \ref{sec:weighted_classification}), the cost-sensitive one-versus-one (CSOVO) SVM modification by Lin is used \cite{Lin2010}.

\section{Experiments} \label{sec:experiments}

{To demonstrate the effectiveness of the transfer learning procedure, the proposed method in its entirety is evaluated, with and without deploying training data from other seats. For  comparative analysis, we show our results against two approaches. The first is the method of Zhang \etal \cite{Zhang2005}\footnote{We developed our own implementation of \cite{Zhang2005} as faithfully as possible.} which relies on a dense Haar feature response as the feature vector and an RBF SVM for classification. Many works use Edge based features instead of Haar responses, such as  \cite{Kong2004,Huang2010,Gao2010}, with variations on the classification stage.  As such, the second method we compare against deploys a standard edge feature extractor (as used in \cite{Huang2010}) followed by PCA and an RBF SVM, and we refer to this hereafter as the ``Classic Approach.''}  
More recent works have used techniques such as finite state machines to combine results from different frames \cite{Huang2010}, whereas here we are focussing on maximising the classification performance on single images.  {We chose the approaches we compare against because clear implementation details were available, they use similar camera positions for the front passenger which indicated the features used would work well with our footage, and the amount of training and testing data we use suggests an SVM is both an appropriate choice of classifier and allows more direct comparison.}

Results are given for the front passenger (FP), rear nearside (RN), rear middle (RM) and rear offside (RO) seats.  The driver's seat is not relevant here, as detection and classification of the driver are unnecessary if the car is being driven. Other than the parameters determined empirically as discussed before (i.e. $\Psi$, $\Phi$ \etc) no other significant parameters are required. 

To reduce the possibility of overfitting to certain occupants, no occupant appears in both the training and testing sets.  To prevent very similar images appearing in either the training or testing sets, images from the same video sequence are chosen that are more than one second apart. The following results are for training and testing sets of 1000 randomly selected images per seat (subject to the above conditions), and the accuracy presented is taken as the mean over 10 runs.

First, results will be presented for standard classifiers to investigate how accurate these methods are at differentiating between all the labelled classes.  This will be followed by results from weighted classifiers tuned to the tasks of occupant detection, classification for child locks and classification for airbag suppression, as would be necessary in a production vehicle.

\subsection{Overall classification performance} \label{sec:overall_classification}

We first test the the classification accuracy over all 8 labelled classes.  These are:
empty seat,
small child in booster seat,
small child in child seat,
small child with no additional seat,
large child in booster seat,
large child in child seat,
large child with no additional seat and
adult.  
In training, children up to Year 2 of school were assigned to the small child classes, and those in Year 3 and above were assigned to the large child classes, making the cutoff around 6 years old.

The outcome of this experiment are displayed in Table \ref{tab:classifiction8}.  The proposed method, using transferred information, averages 78.8\%, an improvement over 76.5\% when not using transferred information, 74.0\% when using the classic approach and 65.9\% when using the method of Zhang \etal \cite{Zhang2005}.

\begin{table}
\begin{center}
\small
\begin{tabular}{l|l|l|l|l||l|}
\cline{2-5}
& \multicolumn{4}{ c| }{\bf Seat} \\ \cline{1-5}
\hline
\multicolumn{1}{ |l|| }{\bf Method} & \bf{FP} & \bf{RN} & \bf{RM} & \bf{RO} & {\bf Avg.}\\
\hline\hline
\multicolumn{1}{ |l|| }{Proposed (non trans.)} & 85.1 & 77.3  & 72.0 & {\bf 71.7} & 76.5\\
\multicolumn{1}{ |l|| }{Proposed} & {\bf 86.0} & {\bf 82.0} & {\bf 77.3} & 69.9 & {\bf 78.8}\\
\multicolumn{1}{ |l|| }{Zhang \cite{Zhang2005}} & 74.7  & 64.3  & 68.0 & 56.5  & 65.9 \\
\multicolumn{1}{ |l|| }{Classic Approach} & 80.5 & 72.5 & 71.8 & 71.0 & 74.0\\
\hline
\end{tabular}
\end{center}
\caption{Unweighted classification percentage accuracy over all eight classes for unweighted classifiers.  FP denotes the front passenger seat, RN the rear nearside seat, RM the rear middle seat and RO the rear offside seat.}
\label{tab:classifiction8}
\end{table}

\begin{table} 
\begin{center}
\small
\begin{tabular}{l|l|l|l|l||l|}
\cline{2-5}
& \multicolumn{4}{ c|| }{\bf Seat} \\ \cline{1-5}
\hline
\multicolumn{1}{ |l|| }{\bf Method} & \bf{FP} & \bf{RN} & \bf{RM} & \bf{RO} & {\bf Avg.}\\
\hline\hline
\multicolumn{1}{ |l|| }{Proposed (non trans.)} & 87.2 & 89.7 & 85.2 & 86.3 & 87.1\\
\multicolumn{1}{ |l|| }{Proposed} & {\bf 90.0} & {\bf 91.2} & {\bf 91.1} & {\bf 88.1} & {\bf 90.1} \\
\multicolumn{1}{ |l|| }{Zhang \cite{Zhang2005}} & 80.8 & 76.8 & 75.6 & 57.8 & 72.7 \\
\multicolumn{1}{ |l|| }{Classic Approach} & 88.1 & 80.1 & 82.0 & 80.6 & 82.7\\
\hline
\end{tabular}
\end{center}
\caption{Unweighted classification percentage accuracy over the empty, adult, small child and large child classes for unweighted classifiers.}
\label{tab:classifiction4}
\end{table}

\subsection{Weighted classification performance}\label{sec:weighted_classification}

For the three use cases of occupant detection, classification for childlocks and classification for airbag suppression, it is possible to reduce the number of labelled classes to four - Empty (E), Adult (A), Small child (S) and Large child (L), with the same child age cutoff as  earlier.  The cost matrices for weighted SVMs used for these tasks are given in Figure \ref{fig:svm_cost}.  Note that the cost matricies chosen for the transfer bias in Section \ref{sec:gaussians} (which feature elements not equal to zero and one to help retain distribution information) do not have to be the same as the cost matrices used for SVM classification.
As a comparison, the unweighed performance over these four superclasses is given in Table \ref{tab:classifiction4}.  The transferred features increase the average classification accuracy in this case from 87.1\% to 90.1\% (the classic approach scores 82.7\% and \cite{Zhang2005} 72.2\%).

Table \ref{tab:detection} shows the weighted classification accuracies, using the cost matrix in Figure \ref{fig:det_mat} for the occupant detection task.  The transferred features increase the average classification accuracy in this case from 93.8\% to 96.7\%, compared to 91.7\% for the classic approach and 85.8\% for \cite{Zhang2005}.  An example decision outcome is shown in Figure \ref{fig:example_detection}.

Next, Table \ref{tab:childlock} shows the results for the child lock task, using the cost matrix in Figure \ref{fig:child_lock_mat}.  This time, the transferred features increase the average classification accuracy from 94.1\% to 97.2\%, compared to 92.2\% for the classic approach and 83.9\% for \cite{Zhang2005}.  An example decision outcome is shown in Figure \ref{fig:example_childlock}.

Finally, Table \ref{tab:airbag} shows the results for the weighted classification for automatic airbag suppression, and uses the cost matrix in Figure \ref{fig:airbag_mat}.  Again, the transferred features increase the average classification accuracy, from 92.0\% to 94.4\%, compared to 89.1\% for the classic approach and 84.3\% for \cite{Zhang2005}.  An example decision outcome is shown in Figure \ref{fig:example_airbag}.

\begin{figure}
    \begin{centering}
    	\subfloat[Occupant detection.\label{fig:det_mat}]{\begin{tabular}{p{1.0mm} | p{1.0mm} p{1.0mm} p{1.0mm} p{1.0mm} }
    	 & E & A & S & L\\
    	\hline
    	E & 0 & 1 & 1 & 1\\
    	A & 1 & 0 & 0 & 0\\
    	S & 1 & 0 & 0 & 0\\
    	L & 1 & 0 & 0 & 0\\
    	\end{tabular}} \hspace{1.8mm}
    	\subfloat[Child locks.\label{fig:child_lock_mat}]{\begin{tabular}{p{1.0mm} | p{1.0mm} p{1.0mm} p{1.0mm} p{1.0mm} }
    	 & E & A & S & L\\
    	\hline
    	E & 0 & 0 & 0 & 0\\
    	A & 0 & 0 & 1 & 1\\
    	S & 1 & 1 & 0 & 0\\
    	L & 1 & 1 & 0 & 0\\
    	\end{tabular}} \hspace{1.8mm}
    	\subfloat[Airbag suppression.\label{fig:airbag_mat}]{\begin{tabular}{p{1.0mm} | p{1.0mm} p{1.0mm} p{1.0mm} p{1.0mm} }
    	 & E & A & S & L\\
    	\hline
    	E & 0 & 0 & 0 & 0\\
    	A & 1 & 0 & 1 & 0\\
    	S & 0 & 1 & 0 & 1\\
    	L & 1 & 0 & 1 & 0\\
    	\end{tabular}}
    \end{centering}
    \caption{Cost matrices used by weighted SVMs.  The classes denoted are Empty seat (E), Adult (A), Small child (S) and Large child (L).}
\label{fig:svm_cost}
\end{figure}

\begin{table} 
\begin{center}
\small
\begin{tabular}{l|l|l|l|l||l|}
\cline{2-5}
& \multicolumn{4}{ c| }{\bf Seat} \\ \cline{1-5}
\hline
\multicolumn{1}{ |l|| }{\bf Method} & \bf{FP} & \bf{RN} & \bf{RM} & \bf{RO} & {\bf Avg.}\\
\hline\hline
\multicolumn{1}{ |l|| }{Proposed (non trans.)} & 96.1 & 93.2 & 93.3 & 92.5 & 93.8  \\
\multicolumn{1}{ |l|| }{Proposed} & {\bf 98.0} & {\bf 95.3} & {\bf 97.4} & {\bf 95.9} & \bf{96.7}  \\
\multicolumn{1}{ |l|| }{Zhang \cite{Zhang2005}} & 90.9  & 88.1 & 89.9 & 74.3  & 85.8 \\
\multicolumn{1}{ |l|| }{Classic Approach} & 96.2 & 90.0 & 90.5 & 90.1 & 91.7  \\
\hline
\end{tabular}
\end{center}
\caption{Weighted classification percentage accuracy for occupant detection using the cost matrix in Figure \ref{fig:det_mat}.}
\label{tab:detection}
\end{table}

\begin{table}
\begin{center}
\small
\begin{tabular}{l|l|l|l|l||l|}
\cline{2-5}
& \multicolumn{4}{ c| }{\bf Seat} \\ \cline{1-5}
\hline
\multicolumn{1}{ |l|| }{\bf Method} & \bf{FP} & \bf{RN} & \bf{RM} & \bf{RO} & {\bf Avg.}\\
\hline\hline
\multicolumn{1}{ |l|| }{Proposed (non trans.)} & 94.2 & 95.1 & 98.2 & 89.2 & 94.1 \\
\multicolumn{1}{ |l|| }{Proposed} & {\bf 98.9} & {\bf 97.1} & {\bf 98.3} & {\bf 94.5} & \bf{97.2} \\
\multicolumn{1}{ |l|| }{Zhang \cite{Zhang2005}} & 90.1 & 88.0 & 87.2  & 70.4  & 83.9 \\
\multicolumn{1}{ |l|| }{Classic Approach} & 94.7 & 91.1 & 92.0 & 90.8 & 92.2 \\
\hline
\end{tabular}
\end{center}
\caption{Weighted classification percentage accuracy for automatic child locks using the cost matrix in Figure \ref{fig:child_lock_mat}.}
\label{tab:childlock}
\end{table}

\begin{table} 
\begin{center}
\small
\begin{tabular}{l|l|l|l|l||l|}
\cline{2-5}
& \multicolumn{4}{ c| }{\bf Seat} \\ \cline{1-5}
\hline
\multicolumn{1}{ |l|| }{\bf Method} & \bf{FP} & \bf{RN} & \bf{RM} & \bf{RO} & {\bf Avg.}\\
\hline\hline
\multicolumn{1}{ |l| }{Proposed (non trans.)} & 92.8 & 93.6 & 92.1 & 89.3 & 92.0 \\
\multicolumn{1}{ |l| }{Proposed} & {\bf 96.2} & {\bf 96.0} & {\bf 94.4} & {\bf 91.1} & \bf{94.4} \\
\multicolumn{1}{ |l|| }{Zhang \cite{Zhang2005}} & 90.4  & 85.4  & 85.7  & 75.6 & 84.3 \\
\multicolumn{1}{ |l| }{Classic Approach} & 92.8 & 88.4 & 87.9 & 87.2 & 89.1 \\
\hline
\end{tabular}
\end{center}
\caption{Weighted classification percentage accuracy for automatic airbag suppression using the cost matrix in Figure \ref{fig:airbag_mat}.}
\label{tab:airbag}
\end{table}

\begin{figure*}
    \begin{centering}
    \subfloat[Occupant detection task.  Green highlighting indicates an occupant is present.\label{fig:example_detection}]{\includegraphics[width=\fisheyewidth\columnwidth]{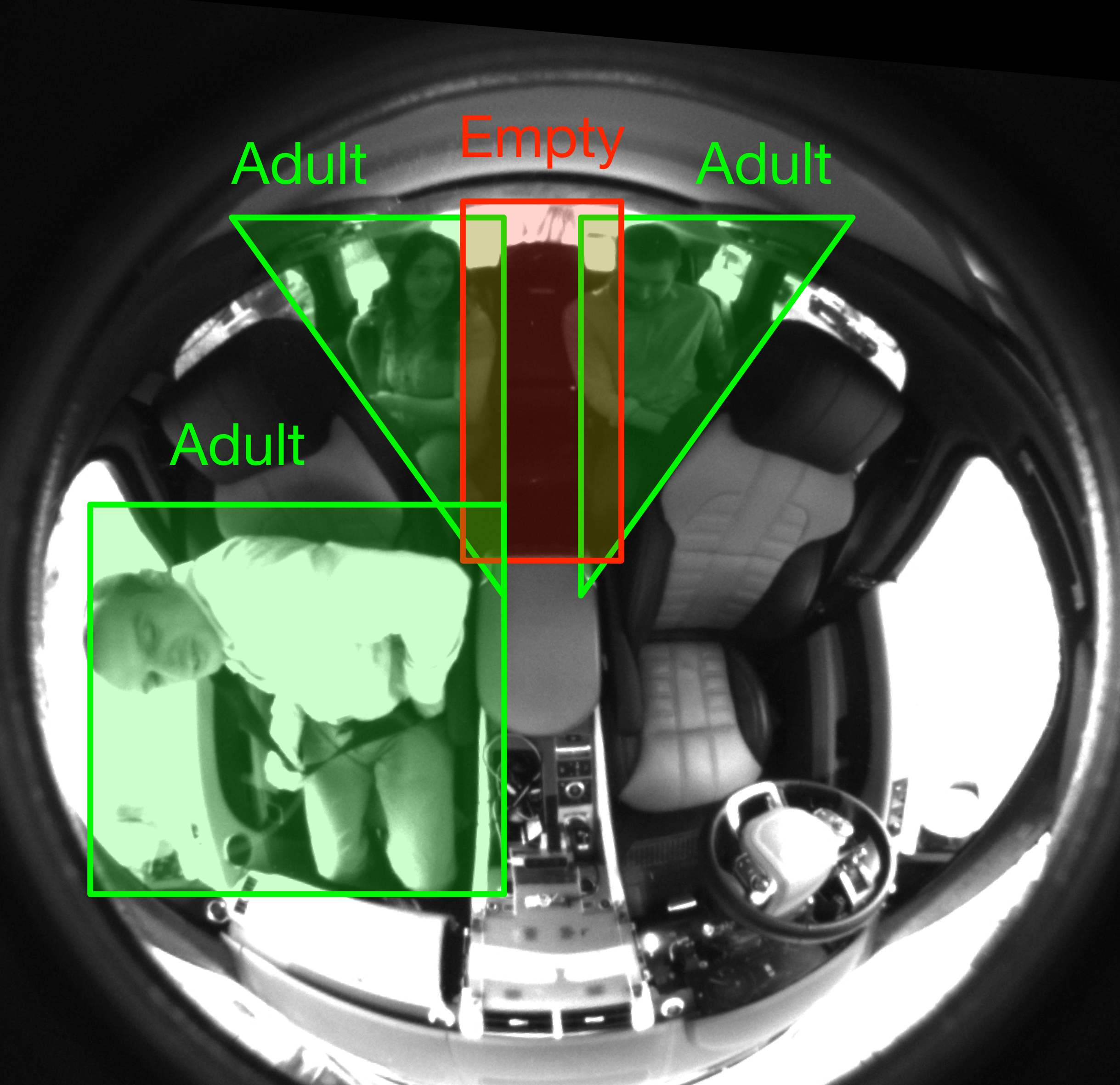}} \hspace{2mm}
    \subfloat[Automatic childlocks.  Green highlighting indicates a childlock should be engaged.\label{fig:example_childlock}]{\includegraphics[width=\fisheyewidth\columnwidth]{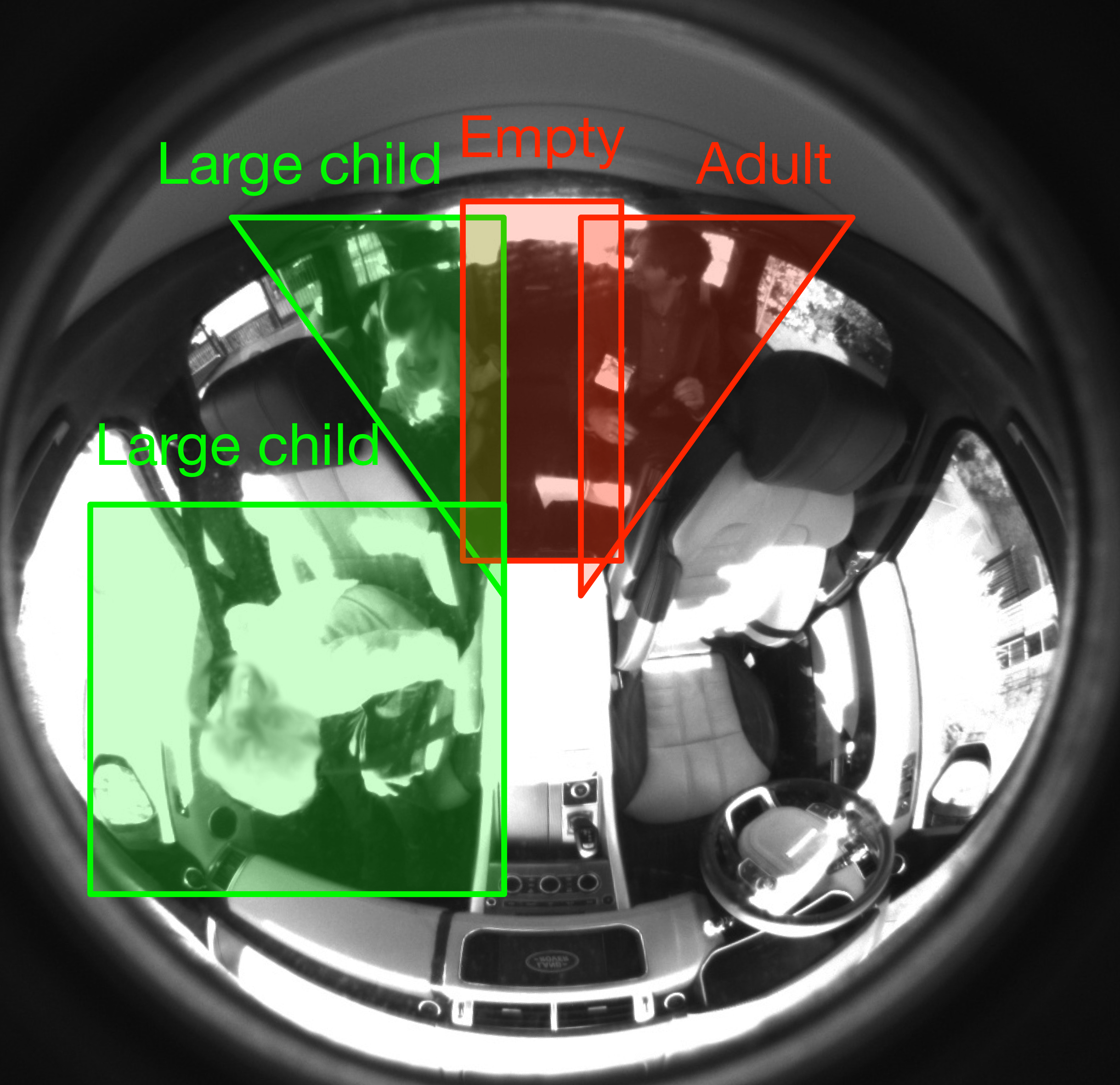}} \hspace{2mm}
    \subfloat[Automatic airbag suppression.  Green highlighting indaces an airbag should be suppressed.\label{fig:example_airbag}]{\includegraphics[width=\fisheyewidth\columnwidth]{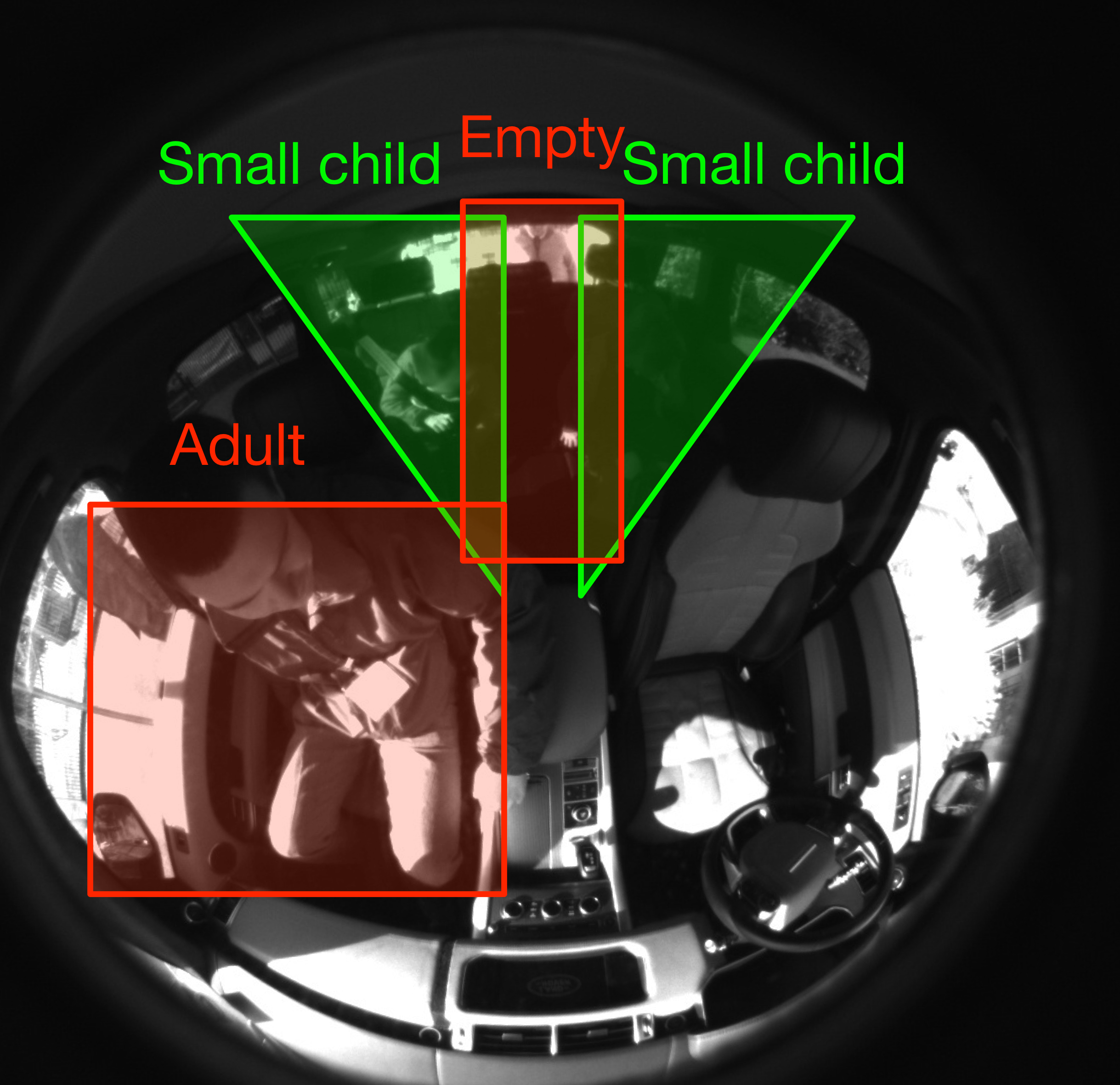}}
    \end{centering}
    \caption{Some example weighted classification decisions with child faces blurred.}
\label{fig:result_examples}
\end{figure*}

\subsection{Discussion}

The most important result to take away from the experiments  is that the inclusion of transferred information does indeed increase the success rate for the weighted classification tasks.  Improvements are found in the successful decisions for occupancy detection, classification for automatic child locks, and classification for automatic airbag suppression.  An improvement is also observed for unweighted classification, although as the proposed method is not designed specifically for this task, it may well be the case that other inductive transfer techniques would perform better for unweighted classification.

Another trend that is visible across all our weighted classification experiments is that the front passenger seat obtains higher scores than  the rear seats.  This is not necessarily surprising, as more detail is visible due to proximity to the camera.  The front seats and occupants can also partially occlude the view of the rear of the vehicle.

An interesting failure case worth noting can be found in Table \ref{tab:classifiction8} which presents the results of the unweighted classification across all classes.  Whilst including transferred features in the training set increases the overall classification accuracy for the front passenger, rear nearside and rear middle seats, including it for the rear offside seat results in a less successful classification rate.  This suggests that the proposed method is more well suited to weighted classification than unweighted and is probably caused by unrelated classes having an equal pull on each feature.

\section{Conclusion}\label{sec:conclusion}

In this paper, a framework was introduced to perform the tasks of occupant detection and occupant classification for automatic child locks and airbag suppression.  The method works for all passenger seats using a single overheard NIR camera.  To make full use of the training data from all seats, a cost-based transfer learning technique was introduced to adjust source features in a reduced space, whilst taking into account the cost of misclassifying these features as certain classes.  An evaluation was performed on a large dataset consisting of 40 adult and 60 child volunteers in real vehicles, and the transferred features contributed to an improved classification accuracy.

There are a number of possible avenues for future work leading on from what has been presented here.  One is to investigate if the proposed method can also incorporate information from standard object datasets \cite{Deng2009} to recognize objects left behind in the cabin, and issue warnings if it is likely to be stolen.  Here, costs can be attributed to certain items - a laptop would have a higher cost of being left behind than a book, for instance.  {Another is to investigate how well features learned in one type of car can be transferred to another type, e.g. those with lower roofs in ``sporty'' models giving a different viewpoint.}

{\small
\bibliographystyle{ieee}
\bibliography{main}

\begin{thebibliography}{10}\itemsep=-1pt

\bibitem{Member2001}
{Technologies , Challenges , and Research and Development Expenditures for
  Advanced Air Bags}.
\newblock {\em Report to the Chairman and Ranking Minority Member, Committee on
  Commerce, Science, and Transportation, U.S. Senate}, 2001.

\bibitem{Alefs2008}
B.~Alefs, M.~Clabian, and M.~Painter.
\newblock {Occupant classification by boosting and PMD-technology}.
\newblock In {\em IEEE Intelligent Vehicles Symposium}, 2008.

\bibitem{Baktashmotlagh2014}
M.~Baktashmotlagh, M.~T. Harandi, B.~C. Lovell, and M.~Salzmann.
\newblock {Domain Adaptation on the Statistical Manifold}.
\newblock In {\em CVPR}, 2014.

\bibitem{SarahBonninThomasH.WeisswangeFranzKummert2014}
S.~Bonnin, T.~H. Weisswange, K.~F., and J.~Schmuedderich.
\newblock {Pedestrian Crossing Prediction using Multiple Context-based Models}.
\newblock In {\em International Conference on Intelligent Transportation
  Systems}, 2014.

\bibitem{Bottazzi2013}
V.~S. Bottazzi, P.~V.~K. Borges, and J.~Jo.
\newblock {A vision-based lane detection system combining appearance
  segmentation and tracking of salient points}.
\newblock {\em IEEE Intelligent Vehicles Symposium}, (4):443--448, 2013.

\bibitem{Brox2004}
T.~Brox, N.~Papenberg, and J.~Weickert.
\newblock {High Accuracy Optical Flow Estimation Based on a Theory for
  Warping}.
\newblock In {\em European Conference on Computer Vision}, 2004.

\bibitem{Campos2013}
T.~Campos, A.~Khan, F.~Yan, N.~Farajidavar, D.~Windridge, J.~Kittler, and
  W.~Christmas.
\newblock {A framework for automatic sports video annotation with anomaly
  detection and transfer learning}.
\newblock In {\em Machine Learning and Cognitive Science}, 2013.

\bibitem{CC01a}
C.~Chang and C.~Lin.
\newblock {\{LIBSVM\}: A library for support vector machines}.
\newblock {\em ACM Transactions on Intelligent Systems and Technology},
  2(27):1--27, 2011.

\bibitem{Cheng2004}
S.~Y. Cheng and M.~M. Trivedi.
\newblock {Human posture estimation using voxel data for "smart" airbag
  systems: issues and framework}.
\newblock In {\em IEEE Intelligent Vehicles Symposium}, 2004.

\bibitem{Cheng2010}
S.~Y. Cheng and M.~M. Trivedi.
\newblock {Vision-based infotainment user determination by hand recognition for
  driver assistance}.
\newblock {\em IEEE Transactions on Intelligent Transportation Systems},
  11(3):759--764, 2010.

\bibitem{Coifman2005}
R.~R. Coifman, S.~Lafon, a.~B. Lee, M.~Maggioni, B.~Nadler, F.~Warner, and
  S.~W. Zucker.
\newblock {Geometric diffusions as a tool for harmonic analysis and structure
  definition of data: diffusion maps.}
\newblock {\em Proceedings of the National Academy of Sciences of the United
  States of America}, 102(21):7426--31, May 2005.

\bibitem{Deng2009}
J.~D.~J. Deng, W.~Dong, R.~Socher, L.~Li, K.~Li, and L.~Fei-Fei.
\newblock {ImageNet: A large-scale hierarchical image database}.
\newblock In {\em CVPR}, 2009.

\bibitem{Devarakota2007}
P.~R. Devarakota.
\newblock {Occupant classification using range images}.
\newblock {\em IEEE Transactions on Vehicular Technology}, 56(4):1983--1993,
  2007.

\bibitem{Farajidavar2014}
N.~Farajidavar.
\newblock {Adaptive Transductive Transfer Machines}.
\newblock In {\em BMVC}, 2014.

\bibitem{Farmer2003}
M.~Farmer and A.~Jain.
\newblock {Occupant classification system for automotive airbag suppression}.
\newblock In {\em CVPR}, 2003.

\bibitem{Farmer2007}
M.~E. Farmer and A.~K. Jain.
\newblock {Smart automotive airbags: Occupant classification and tracking}.
\newblock {\em IEEE Transactions on Vehicular Technology}, 56(1):60--80, 2007.

\bibitem{Gao2010}
Z.~Gao and L.~Duan.
\newblock {Vision detection of vehicle occupant classification with Legendre
  moments and Support Vector Machine}.
\newblock In {\em IEEE International Congress on Image and Signal Processing},
  2010.

\bibitem{Garcia2012}
I.~Garcia, S.~Bronte, L.~M. Bergasa, J.~Almazan, and J.~Yebes.
\newblock {Vision-based drowsiness detector for real driving conditions}.
\newblock In {\em IEEE Intelligent Vehicles Symposium}, 2012.

\bibitem{George2010}
B.~George, H.~Zangl, T.~Bretterklieber, and G.~Brasseur.
\newblock {A combined inductivecapacitive proximity sensor for seat occupancy
  detection}.
\newblock {\em IEEE Transactions on Instrumentation and Measurement},
  59(5):1463--1470, 2010.

\bibitem{Glass2000}
R.~J. Glass, M.~Segui-Gomez, and J.~D. Graham.
\newblock {Child passenger safety: decisions about seating location, airbag
  exposure, and restraint use.}
\newblock 20(4):521--7, 2000.

\bibitem{Goktuk2005}
S.~Goktuk and A.~Rafii.
\newblock {An occupant classification system eigen shapes or knowledge-based
  features}.
\newblock In {\em CVPR}, 2005.

\bibitem{Gretton2012}
A.~Gretton.
\newblock {A Kernel Two-Sample Test}.
\newblock {\em Journal of Machine Learning Research}, 13:723--773, 2012.

\bibitem{Hanwell2012}
D.~Hanwell and M.~Mirmehdi.
\newblock {Detection of Lane Departure on High-speed Roads.}
\newblock In {\em International Conference on Pattern Recognition Applications
  and Methods}, 2012.

\bibitem{Huang2011}
S.~Huang, E.~Jian, and P.~Hsiao.
\newblock {Occupant classification invariant to seat movement for smart
  airbag}.
\newblock In {\em IEEE International Conference on Vehicular Electronics and
  Safety}, 2011.

\bibitem{Huang2012}
S.~S. Huang.
\newblock Discriminatively trained patch-based model for occupant
  classification.
\newblock {\em IET Intelligent Transport Systems}, 6(2):132--138, 2012.

\bibitem{Huang2010}
S.-S. Huang and P.-Y. Hsiao.
\newblock {Occupant classification for smart airbag using Bayesian filtering}.
\newblock In {\em International Conference on Green Circuits and Systems},
  2010.

\bibitem{Kim2013}
B.~Kim and J.~Pineau.
\newblock {Maximum Mean Discrepancy Imitation Learning}.
\newblock In {\em Robotics: Science and Systems}, 2013.

\bibitem{Koch2002}
C.~Koch, T.~Ellis, and A.~Georgiadis.
\newblock {Real-time occupant classification in high dynamic range
  environments}.
\newblock In {\em IEEE Intelligent Vehicles Symposium}, 2002.

\bibitem{Kong2004}
H.~Kong, Q.~Sun, W.~Bauson, S.~Kiselewich, P.~Ainslie, and R.~Hammoud.
\newblock {Disparity Based Image Segmentation For Occupant Classification}.
\newblock In {\em CVPR Workshop}, 2004.

\bibitem{Lagarias1998}
J.~C. Lagarias, J.~Reeds, M.~H. Wright, and P.~E. Wright.
\newblock {Convergence Properties of the Nelder--Mead Simplex Method in Low
  Dimensions}.
\newblock {\em SIAM Journal on Optimization}, 9(1):112--147, 1998.

\bibitem{Lin2010}
H.-T. Lin.
\newblock In {\em National Taiwan University, Tech. Rep}, 2010.

\bibitem{Long2013}
M.~Long, J.~Wang, G.~Ding, J.~Sun, and P.~S. Yu.
\newblock {Transfer Feature Learning with Joint Distribution Adaptation}.
\newblock In {\em International Conference on Computer Vision}, 2013.

\bibitem{mehney2000vehicle}
M.~A. Mehney, M.~C. McCarthy, M.~G. Fullerton, and F.~J. Malecke.
\newblock {Vehicle occupant weight sensor apparatus}, 2000.

\bibitem{Monwar2013}
M.~M. Monwar and B.~V.~K. Vijaya~Kumar.
\newblock {Vision-based potential collision detection for reversing vehicle}.
\newblock {\em IEEE Intelligent Vehicles Symposium}, (4):88--93, 2013.

\bibitem{Pan2008}
S.~J. Pan, J.~T. Kwok, and Q.~Yang.
\newblock {Transfer Learning via Dimensionality Reduction}.
\newblock In {\em AAAI Conference on Artificial intelligence}, 2008.

\bibitem{Pan2010a}
Z.~Pan, Y.~Li, M.~Zhang, C.~Sun, K.~Guo, X.~Tang, and S.~Z. Zhou.
\newblock In {\em IEEE Virtual Reality Conference}, 2010.

\bibitem{Perronnin2006}
F.~Perronnin and C.~Dance.
\newblock {Fisher Kenrels on Visual Vocabularies for Image Categorizaton}.
\newblock In {\em CVPR}, 2006.

\bibitem{Rohrbach2013}
M.~Rohrbach, S.~Ebert, and B.~Schiele.
\newblock {Transfer learning in a transductive setting}.
\newblock In {\em Neural Information Processing Systems}, 2013.

\bibitem{Scaramuzza2006}
D.~Scaramuzza, A.~Martinelli, and R.~Siegwart.
\newblock {A toolbox for easily calibrating omnidirectional cameras}.
\newblock In {\em IEEE International Conference on Intelligent Robots and
  Systems}, 2006.

\bibitem{seip2002linear}
R.~Seip.
\newblock {Linear ultrasound transducer array for an automotive occupancy
  sensor system}, 2002.

\bibitem{Shao2012}
H.~Shao, B.~Tong, and E.~Suzuki.
\newblock {Extended MDL principle for feature-based inductive transfer
  learning}.
\newblock {\em Knowledge and Information Systems}, 35(2):365--389, 2012.

\bibitem{Sivaraman2013}
S.~Sivaraman and M.~M. Trivedi.
\newblock {Looking at Vehicles on the Road: A Survey of Vision-Based Vehicle
  Detection, Tracking, and Behavior Analysis}.
\newblock {\em IEEE Transactions on Intelligent Transportation Systems},
  14(4):1773--1795, 2013.

\bibitem{Vicente2015}
F.~Vicente, Z.~Huang, X.~Xiong, F.~Torre, W.~Zhang, and D.~Levi.
\newblock {Driver Gaze Tracking and Eyes Off the Road Detection System}.
\newblock {\em IEEE Transactions on Intelligent Transportation Systems},
  16(4):1--14, 2015.

\bibitem{wallace2003vehicle}
M.~W. Wallace and A.~M.~W. Wallace.
\newblock {Vehicle occupant classification system and method}, 2003.

\bibitem{Wang2014}
S.~Wang.
\newblock {A New Transfer Learning Boosting Approach Based on Distribution
  Measure with an Application on Facial Expression Recognition}.
\newblock In {\em International Joint Conference on Neural Networks}, 2014.

\bibitem{Zhang2005}
Y.~Zhang, S.~Kiselewich, and W.~Bauson.
\newblock {A monocular vision-based occupant classification approach for smart
  airbag deployment}.
\newblock In {\em IEEE Intelligent Vehicles Symposium}, 2005.

\end{thebibliography}
}

\end{document}